\documentclass[10pt,journal,compsoc]{IEEEtran}
\usepackage{cite}
\usepackage[pdftex]{graphicx}
\usepackage{array}
\usepackage{bbm}
\usepackage{amsmath}
\usepackage{amssymb}
\usepackage{enumitem}
\usepackage{subcaption}
\usepackage{bbding}

\usepackage[pagebackref=false,breaklinks=true,bookmarks=false]{hyperref}

\newcolumntype{P}[1]{>{\centering\arraybackslash}p{#1}}

\usepackage{multirow,booktabs}
\usepackage{algorithm}
\usepackage{algpseudocode}
\usepackage{pifont}
\usepackage{colortbl}
\usepackage{xcolor}
\newcommand{\xmark}{\ding{55}}
\newcommand{\cmark}{\ding{51}}
\newcommand{\et}{\textit{et al.\ }}

\definecolor{lightorange}{rgb}{0.8, 0.4, 0.0}

\definecolor{darkgray}{rgb}{0.8, 0.8, 0.8}
\algrenewcommand\algorithmicrequire{\textbf{Input:}}
\algrenewcommand\algorithmicensure{\textbf{Output:}}

\begin{document}
\title{Vid-Morp: Video Moment Retrieval Pretraining\\from Unlabeled Videos in the Wild}

\author{
Peijun Bao,
\and
Chenqi Kong,
\and
Zihao Shao,
\and
Boon Poh Ng,\\
\and
Meng Hwa Er,~\IEEEmembership{Life~Fellow,~IEEE},
\and
Alex C. Kot,~\IEEEmembership{Life~Fellow,~IEEE}
\IEEEcompsocitemizethanks{
\IEEEcompsocthanksitem 
Nanyang Technological University, Singapore.
%
\IEEEcompsocthanksitem{
E-mail: peijun001@e.ntu.edu.sg
}
}
}

\markboth{}%
{Shell \MakeLowercase{\textit{et al.}}: Bare Demo of IEEEtran.cls for Computer Society Journals}

\IEEEtitleabstractindextext{%
\begin{abstract}
Given a natural language query, video moment retrieval aims to localize the described temporal moment in an untrimmed video.
A major challenge of this task is its heavy dependence on labor-intensive annotations for training.
Unlike existing works that directly train models on manually curated data, we propose a novel paradigm to reduce annotation costs: pretraining the model on unlabeled, real-world videos.
To support this, we introduce \textbf{Vid}eo \textbf{M}oment \textbf{R}etrieval \textbf{P}retraining (\textbf{Vid-Morp}), a large-scale dataset collected with minimal human intervention, consisting of over 50K videos captured in the wild and 200K pseudo annotations.
Direct pretraining on these imperfect pseudo annotations, however, presents significant challenges, including mismatched sentence-video pairs and imprecise temporal boundaries.
To address these issues, we propose the ReCorrect algorithm, which comprises two main phases: semantics-guided refinement and memory-consensus correction.
The semantics-guided refinement enhances the pseudo labels by leveraging semantic similarity with video frames to clean out unpaired data and make initial adjustments to temporal boundaries.
In the following memory-consensus correction phase, a memory bank tracks the model predictions, progressively correcting the temporal boundaries based on consensus within the memory.
Comprehensive experiments demonstrate ReCorrect's strong generalization abilities across multiple downstream settings.
\textbf{Zero-shot ReCorrect} achieves over \textbf{75\%} and \textbf{80\%} of the best fully-supervised performance on two benchmarks, while \textbf{unsupervised ReCorrect} reaches about \textbf{85\%} on both.
The code, dataset, and pretrained models are available at \href{https://github.com/baopj/Vid-Morp}{https://github.com/baopj/Vid-Morp}.
\end{abstract}
}

\maketitle

\IEEEdisplaynontitleabstractindextext

\ifCLASSOPTIONpeerreview
\begin{center} \bfseries EDICS Category: 3-BBND \end{center}
\fi

\IEEEpeerreviewmaketitle

\IEEEraisesectionheading{\section{Introduction}\label{sec:introduction}}
%
\IEEEPARstart{G}{iven} a natural language query and an untrimmed video, the task of Video Moment Retrieval (VMR)~\cite{tall,dense_cap} aims to temporally localize the video moment described by the language query.
VMR is one of the most fundamental tasks in video understanding and has a wide range of real-world applications~\cite{Qi2021SemanticsAwareSB,Sreenu2019IntelligentVS,Jin2024Weak,Zhu2021DSNetAF}, such as video summarization, robot manipulation,  and video surveillance analysis.

In recent years, the performance of VMR has been improved by deep learning techniques~\cite{attention_2,Mun2020LocalGlobalVI,cbp,man,Cai2024TemporalSG,Bao2024SimBase,Zhang2020Learning2T,Zhang2019CrossModalIN,Bao2022LearningSI,Bao2021DenseEG,bao2024omnipotent,bao2024local} and the availability of manually annotated data~\cite{tall,dense_cap}.
However, collecting these annotations, including sentence queries and temporal boundaries, remains expensive, labor-intensive, and not scalable.
Additionally, such annotations often exhibit language and temporal biases~\cite{Bao2023CrossModalLC,Yuan2021ACL,Li2022CompositionalTG} (\textit{e.g.} biases in query style and temporal boundary distribution), limiting their practical applicability.

To tackle this challenge, recent studies~\cite{Nam2021ZeroshotNL,Zheng2023GeneratingSP,Kim2022LanguagefreeTF} shift focus to unlabeled videos, exploring unsupervised learning.
These methods, however, share a common limitation: they depend on unlabeled videos drawn from well-annotated datasets~\cite{tall,dense_cap,hendricks17iccv}.
This dependency inherently introduces human labor, as annotators must manually pre-clean these videos, which restricts the scalability of these methods.
Moreover, the training and testing videos in these studies often share a similar distribution, a condition rarely encountered in practical, real-world applications.
The potential for leveraging entirely unlabeled videos from diverse, in-the-wild environments remains largely unexplored.

\begin{figure}[t!]
\centering
\includegraphics[width=0.9\linewidth]{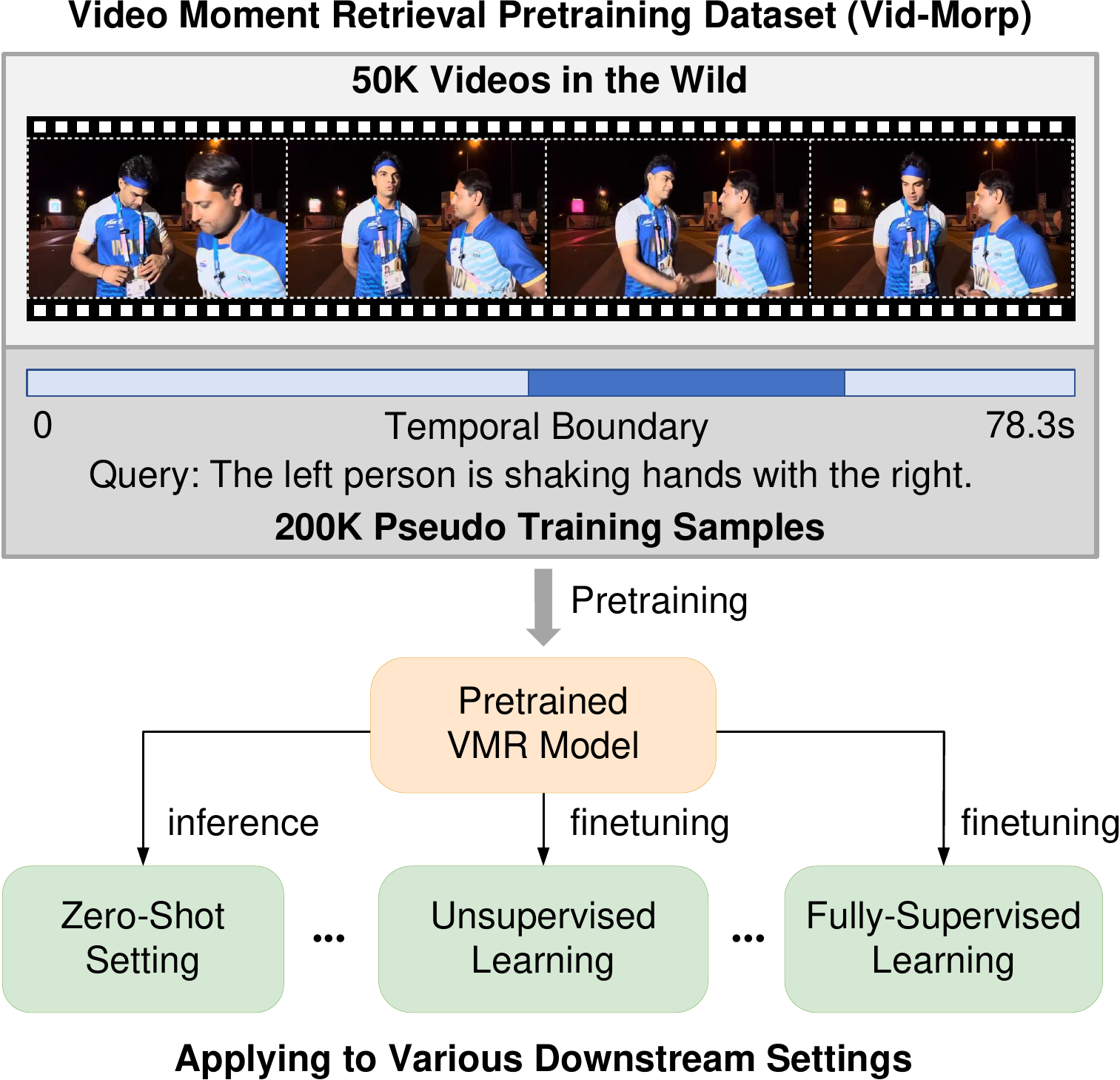}
\caption{
A crucial challenge in video moment retrieval is its heavy reliance on extensive manual annotations for training.
To overcome this, we introduce a large scale dataset for Video Moment Retrieval Pretraining (Vid-Morp), collected with minimal human involvement.
Vid-Morp comprises over 50K in-the-wild videos and 200K pseudo training samples.
Models pretrained on Vid-Morp significantly relieve the annotation costs and demonstrate strong generalizability across diverse downstream settings.
}\label{fig_motivation}
\end{figure}

%
To this end, as shown in Fig.~\ref{fig_motivation}, we introduce \textbf{Vid}eo \textbf{M}oment \textbf{R}etrieval \textbf{P}retraining (\textbf{Vid-Morp}), a large-scale dataset comprising over 50K untrimmed videos captured in the wild.
By utilizing multimodal large language models like GPT-4o, we craft tailored prompts to generate pseudo-annotations, resulting in over 200K training samples specifically designed for video moment retrieval.
However, due to the minimal human involvement in creating these samples, directly pretraining on them poses significant challenges.
Common issues in these samples include videos with minimal meaningful activity, mismatched video-query pairs, and imprecise temporal boundaries.

To address these issues, we propose the \textbf{Re}finement and \textbf{Correct}ion (\textbf{ReCorrect}) algorithm, which consists of two main phases: semantics-guided refinement and memory-consensus correction.
In the semantics-guided refinement phase, we enhance the pseudo labels by leveraging semantic similarity between video frames and pseudo labels to clean error-prone training samples, such as idle videos and mismatched video-query pairs, while initially adjusting the temporal boundaries.
In the subsequent memory-consensus correction phase, a memory bank continuously tracks the model's predictions during pretraining.
This memory bank then serves as a reference to progressively calibrate the temporal boundaries of the pseudo labels based on consensus within the memory.

As illustrated in Fig.~\ref{fig_motivation}, the pretrained ReCorrect model can be seamlessly adapted to various downstream settings for VMR, including zero-shot inference, unsupervised and fully supervised learning, as well as out-of-distribution scenarios.
Experiments demonstrate that \textbf{unsupervised ReCorrect} attains about \textbf{85\%} of the state-of-the-art fully supervised performance on both benchmarks of Charades-STA~\cite{tall} and ActivityNet Captions~\cite{dense_cap}.
And \textbf{zero-shot ReCorrect} surpasses \textbf{75\%} and \textbf{80\%} on them respectively.
This highlights the Vid-Morp's potential to address the critical challenge of heavy reliance on manual annotations in VMR.
ReCorrect also effectively mitigates the \textbf{annotation bias} problem~\cite{Yuan2021ACL,Li2022CompositionalTG}, with its zero-shot version outperforming fully supervised methods tailored for this issue across two out-of-distribution benchmarks.
%

%
Our main contributions can be summarised as follows:
\begin{enumerate}
\item
We introduce Vid-Morp, a large scale and diverse dataset containing over 50K  videos captured in the wild and 200K pseudo training samples designed for pretraining in video moment retrieval.
\item
To tackle the issues of error-prone pseudo training samples, we propose the ReCorrect algorithm.
ReCorrect incorporates semantics guided refinement to clean and adjust pseudo labels and exploit memory consensus correction to calibrate  temporal boundaries based on consensus within a memory bank.
\item
Comprehensive experiments demonstrate ReCorrect's state-of-the-art performance across various settings, including zero-shot, unsupervised, fully supervised learning, and out-of-distribution scenarios.
\end{enumerate}

\begin{table*}[t!]
\centering 
\caption{ 
Statistics of video moment retrieval datasets, with relevant metrics reported to compare our dataset against others on training split.
}
\label{table_data_compare} \scalebox{1.0}{ %
\begin{tabular}{l||c||r|r||r|r|r|r|r|r}
\hline 
\multirow{3}{*}{Dataset } & \multirow{3}{*}{%
\begin{tabular}{c}
Labor\\
Free\\
\end{tabular}} & \multicolumn{2}{c||}{Videos} & \multicolumn{6}{c}{Language Queries}\tabularnewline
\cline{3-10}
 &  & Video  & Total  & Query  & Total  & \multicolumn{4}{c}{Vocabulary}\tabularnewline
\cline{7-10}
 &  & Numbers  & Duration  & Numbers  & Tokens  & \multicolumn{1}{c|}{Adj.} & \multicolumn{1}{c|}{Nouns} & \multicolumn{1}{c|}{Verbs} & \multicolumn{1}{c}{Total}\tabularnewline
\midrule            TACoS~\cite{tacos} & \xmark &           0.1K &             4.7h &            9.8K &          0.1M &          0.3K &           0.8K &          0.7K &           1.7K \\
      Charades-STA~\cite{tall} & \xmark &           5.3K &            45.8h &           12.4K &          0.8M &          0.1K &           0.6K &          0.4K &           1.1K \\
 DiDeMO~\cite{hendricks17iccv} & \xmark &           8.5K &            70.9h &           33.0K &          0.2M &          1.3K &           4.8K &          2.4K &           7.0K \\
Anet-Captions~\cite{dense_cap} & \xmark &          10.0K &           326.1h &           37.4K &          0.5M &          2.1K &           7.9K &          4.0K &          11.7K \\
\midrule

      \textbf{Vid-Morp (Ours)} & \cmark & \textbf{52.7K} & \textbf{1013.2h} & \textbf{200.3K} & \textbf{2.1M} & \textbf{6.5K} & \textbf{21.0K} & \textbf{6.6K} & \textbf{31.1K} \\\bottomrule
\end{tabular}
}
\end{table*}
\begin{figure*}[t!]
\centering
\includegraphics[width=0.84\linewidth]{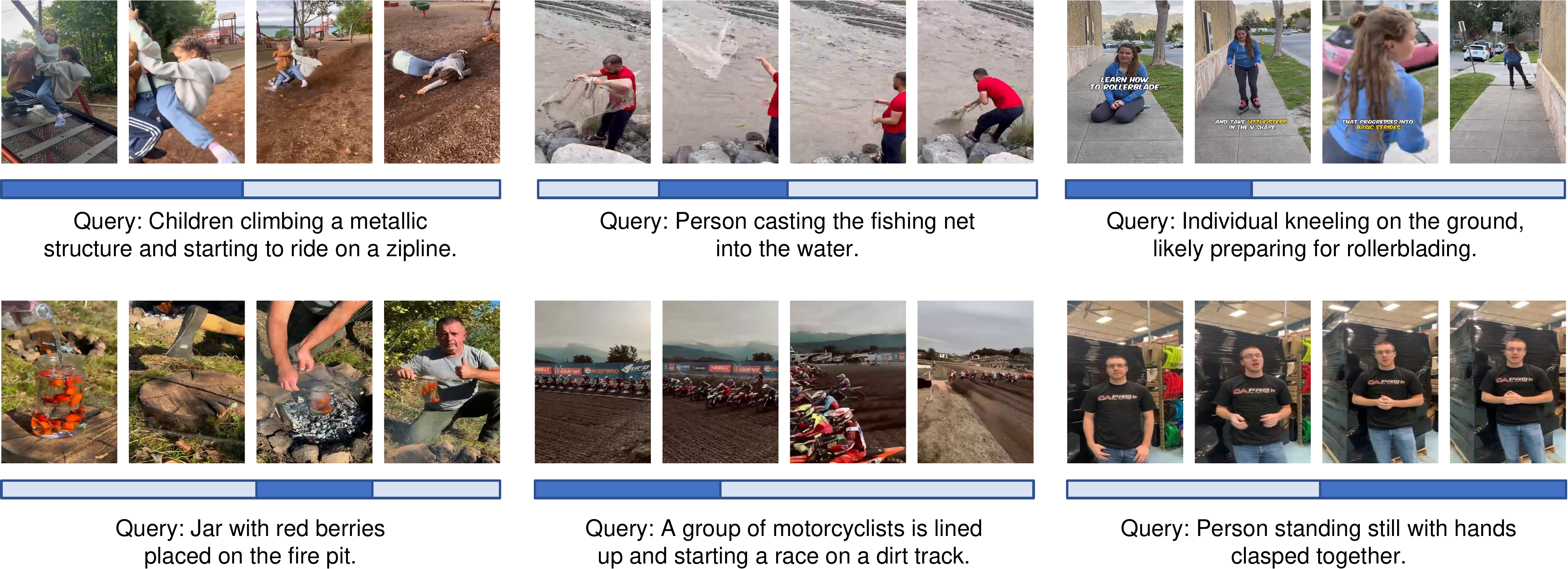}
\caption{
Illustration of video samples and pseudo-annotations, including sentence queries and temporal boundaries, from the Video Moment Retrieval Pretraining (Vid-Morp) dataset.
The dark blue box represents the temporal boundary of the described video moment.
}\label{fig_benchmark}
\end{figure*}

\section{Related Works}
\label{sec_related_works}
\noindent\textbf{Fully-Supervised Video Moment Retrieval.}
The performance of fully supervised Video Moment Retrieval (VMR) has been improved by the advancement of deep learning techniques~\cite{attention_2,Mun2020LocalGlobalVI,Bao2021DenseEG,cbp,man,Zhang2020Learning2T,Zhang2019CrossModalIN} and the availability of manually annotated data~\cite{tall,dense_cap}.
For instance,
Liu \et~\cite{attention_2} advise applying attention mechanism to highlight the crucial part of visual features.
Bao \et~\cite{Bao2021DenseEG} develop an event propagation network to retrieve video moments that are semantically related and temporally coordinated.
The compositional property of query sentences is utilized in~\cite{MM20,comp,Zhang2019CrossModalIN} for temporal reasoning.
While achieving promising performance, these fully-supervised methods rely on the manual annotations  which are labor-intensive and subjective to label.

\noindent\textbf{Unsupervised Video Moment Retrieval.}
To eliminate high cost of annotation costs, unsupervised learning has drawn attention in recent years in various tasks~\cite{bao2024e3m,Bao2023CrossModalLC,gong2020learning,soomro2017unsupervised,Liao2024VideoINSTAZL} of video understanding.
Some recent works~\cite{Nam2021ZeroshotNL,Zheng2023GeneratingSP,Kim2022LanguagefreeTF} investigate unsupervised VMR using only unlabeled videos.
For example, Kim \et~\cite{Kim2022LanguagefreeTF} propose language-free training algorithm to train the VMR model without language data.
However, a major limitation of these models is the unavoidable introduction of manual curation, as they rely on clean videos from existing manually annotated datasets.
Scaling these models to handle real-world video scenarios, which often include  noisy data such as idle videos with no meaningful activities, remains a challenge.
In contrast, the proposed ReCorrect is specifically designed to rely solely on unlabeled videos captured in the wild.

\noindent\textbf{Video Moment Retrieval Pretraining.}
The closest work to ours is ProTeGe~\cite{Wang2023ProTGUP}, which pretrains the feature extraction backbone for VMR.
However, our work differs and complements theirs: while their approach focuses on pretraining feature extraction backbone, ours targets pretraining the retrieval model with a fixed backbone.
This distinction is further emphasized by the fact that our model supports zero-shot setting without additional finetuning, whereas theirs cannot.

UniTVG~\cite{Lin2023UniVTGTU} introduces a general-purpose pretraining dataset for various video-language tasks.
However, its VMR performance is significantly lower than other zero-shot approaches tailored to VMR, reaching only half the metrics (see Table~\ref{table_zero_shot}).
This indicates the necessity of designing pretraining datasets specifically  for VMR.

\section{Vid-Morp Dataset}
\subsection{Overview}
Video moment retrieval (VMR)~\cite{tall,dense_cap} aims to temporally identify the video moment in an untrimmed video as described by a language query.
Although fully-supervised methods achieve promising performance, the high cost of annotation still limits the practical application of VMR.
While recent studies~\cite{Nam2021ZeroshotNL,Zheng2023GeneratingSP,Kim2022LanguagefreeTF} explore unsupervised settings, they continue to rely on clean videos drawn from well-annotated datasets. This dependence introduces manual intervention, making these approaches impractical for real-world scenarios.
The potential for leveraging purely unlabeled videos from diverse, in-the-wild environments remains largely unexplored.

To this end, as presented in Fig~\ref{fig_benchmark}, we introduce Video Moment Retrieval Pretraining (Vid-Morp), a large-scale dataset containing over 50K videos captured in the wild and 200K training annotations, collected with minimal human intervention.
As summarized in Table~\ref{table_data_compare}, our dataset contains five times the number of videos and queries compared to the previous dataset, ActivityNet Captions~\cite{dense_cap}, and encompasses a wealth of semantic content, including diverse activities across various visual domains.

\subsection{Dataset Construction}
To collect the videos, we define a list of target activities and use web crawling to gather videos up to a maximum duration of $t_\text{max}$.
Each video is uniformly sampled into $n_\text{v2f}$ frames, which are then concatenated into a single image.
We employ a multimodal language model (MLLM), such as GPT-4o, to generate pseudo labels, guided by a carefully designed prompt that instructs the MLLM to produce descriptive sentences with associated frame indices matching the image content.
These frame indices are then mapped to start and end timestamps within each video.
This process is designed to be highly scalable, with minimal manual intervention primarily limited to defining search keywords and designing the prompt.

\begin{figure*}[t!]
\centering
\includegraphics[width=1\linewidth]{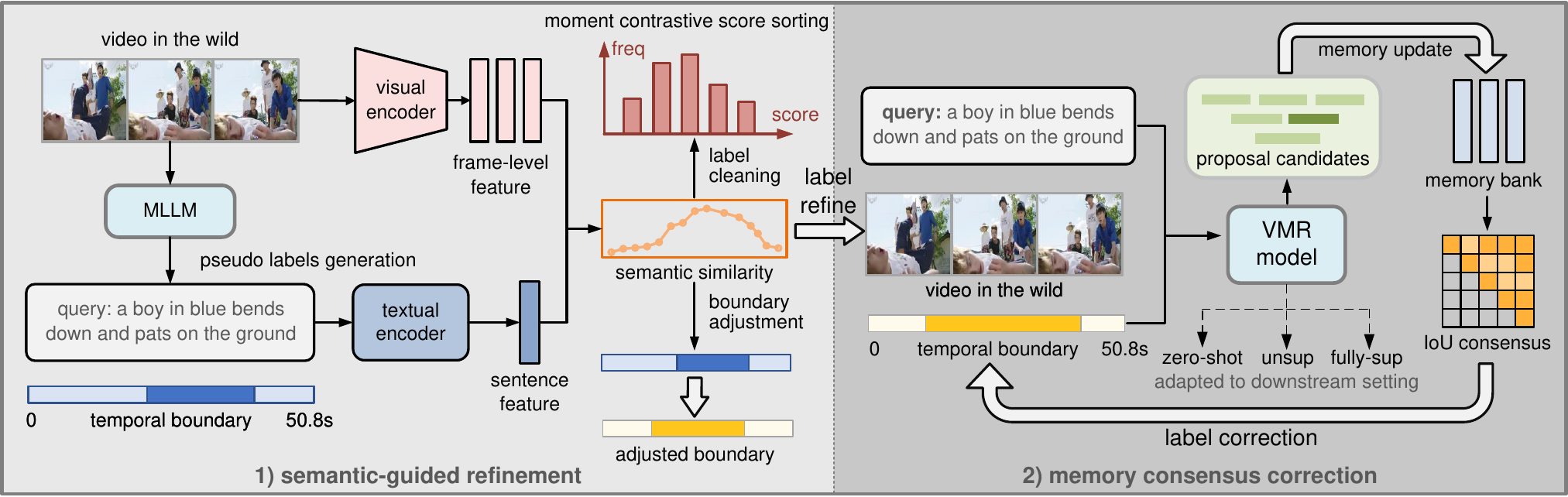}
\caption{
Overview of the Refinement and Correction (ReCorrect) algorithm for video moment retrieval pretraining from in-the-wild videos.
ReCorrect consists of two key phases:
1) semantics-guided refinement, which leverages semantic similarity to clean noisy psuedo training samples, such as idle videos and unmatched video-query pairs, while initially adjusting temporal boundaries,
and 2) memory-consensus correction, where a memory bank tracks model predictions, progressively correcting temporal boundaries based on consensus within the memory.
}\label{fig_method}
\end{figure*}

\begin{figure*}[t!]
\centering
\includegraphics[width=1.0\linewidth]{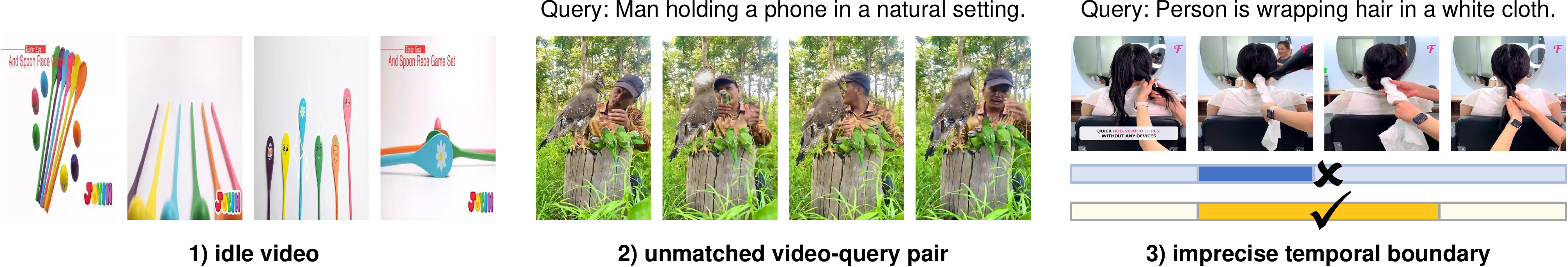}
\caption{
Collected in a scalable, labor-free manner, the Vid-Morp dataset exhibits three common errors in pseudo training samples:
1) idle videos lacking meaningful activity,
2) unmatched video-query pairs where the query event does not appear from the video, and
3) imprecise temporal boundaries where video-query matches are correct but temporal boundaries are inaccurate.
}\label{fig_benchmark_challenge}
\end{figure*}

\section{ReCorrect Algorithm}
Since videos captured in the wild are inherently unclean, and MLLM's limitations in labeling accuracy introduce additional noise, there are widespread errors in the pseudo annotations of the Vid-Morp Dataset, as shown in Fig.~\ref{fig_benchmark_challenge}.
These errors fall into three main categories:
1) Idle videos that lack any meaningful events.
2) Unmatched video-query pairs where the pseudo queries do not correspond to any video frames.
3) Imprecise temporal boundaries where the query matches the video, but the temporal alignment is inaccurate.
These errors present significant challenges for direct pretraining on the pseudo training samples.

To deal with these issues, as illustrated in Fig.~\ref{fig_method}, we propose the Refinement and Correction (ReCorrect) algorithm.
It consists of 1) semantics-guided refinement, which removes erroneous training samples and initially adjusts temporal boundaries, and 2) memory-consensus correction, where a memory bank tracks predictions to correct boundaries based on consensus.

\subsection{Pretraining on Vid-Morp}
\subsubsection{Semantics-Guided Refinement}
%
To ensure that the pseduo query $Q$ are aligned with the video moment in the untrimmed video, we propose a semantics guided refinement to clean out the unmatched pair of video and sentence, as well as adjust the pseudo temporal boundary.
We first extract the query feature $q$ and the visual feature $v_t$ for $t$-th frame with pretrained CLIP model~\cite{Radford2021LearningTV}, and compute the semantic similarities $s_t$ between them, formulated as:
\begin{align}
s_t &= \frac{{q^\top v_t}}{{||q|| \cdot ||v_t||}}, \quad t = 1 \ldots T
\label{eq_instance_sim_step2}
\end{align}
where $T$ is the total number of frames in the video.

Let $b=(\tau_s, \tau_e)$ denote the pseudo-temporal boundaries provided by the MLLM for the query, where $\tau_s$ and $\tau_e$ represent the start and end timepoints, respectively.
Then, we compute the moment contrastive scores, which indicate the contrastive semantic relevance of the video content to the sentence, comparing the content inside the pseudo-temporal moment versus outside it, formulated as:
\begin{align}
\gamma (\tau_s, \tau_e) = \frac{\sum^{\tau_e}_{t=\tau_s} s_t}{\sum_{t=1}^{\tau_s-1} s_t + \sum_{t=\tau_t+1}^{T} s_t}
\end{align}
A high value of $\gamma (\tau_s, \tau_e)$ indicates strong relevance between the pseudo query and the video moment defined by the temporal boundaries.
We sort the moment contrastive score $\gamma$ for each data sample in descending order, cleaning out the bottom $R$ percent and selecting only the remaining as training samples.

Subsequently, we adjust the pseudo temporal boundary by either shrinking or expanding the start time $\tau_s$ based on the semantic similarity $s_t$.
Specifically, if $\gamma (\tau_s, \tau_e) < \alpha_1 \cdot \gamma (\tau_s^\prime, \tau_s)$, we shrink $\tau_s$ by $\delta$, updating $\tau_s^\prime$ as $\tau_s^\prime = \tau_s - \delta$.
Otherwise, if $\gamma (\tau_s, \tau_e) < \alpha_2 \cdot \gamma (\tau_s^\prime, \tau_s)$, we expand $\tau_s$ by $\delta$, assigning $\tau_s^\prime$ as $\tau_s^\prime = \tau_s + \delta$,
where $\alpha_1$ and $\alpha_2$ are predefined hyperparameters.
This process can be formulated as:
\begin{equation}
\tau_s^\prime =
\begin{cases}
 \tau_s - \delta, & \text{if } \gamma (\tau_s, \tau_e) < \alpha_1 \cdot \gamma (\tau_s^\prime, \tau_s) \\
 \tau_s + \delta, & \text{elif } \gamma (\tau_s, \tau_e) < \alpha_2 \cdot \gamma (\tau_s^\prime, \tau_s) \\
\end{cases}
\end{equation}
We repeat this process until no further adjustments are made to $\tau_s$.
The same approach is also applied to $\tau_e$ to refine the end time point.
The final adjusted pseudo-temporal boundary is denoted as $\hat{b}$.

\subsubsection{Memory Consensus Correction}
%
Although the pseudo temporal boundaries are initially improved through semantics guided refinement, they remain inaccurate and may not fully align with the sentence queries.
To address this, we introduce a memory consensus correction method that calibrates the boundaries in a coarse-to-fine manner.
We maintain a memory bank $\mathcal{M}$ to store potential candidates for pseudo temporal boundaries.
For the $i$-th data sample, its memory bank $\mathcal{M}_i$ is initialized as $\{\hat{b}_i\}$, where $\hat{b}_i$ denotes the temporal boundaries adjusted by the semantics-guided refinement.

We use the same model architecture as the fully supervised VMR model SimBase~\cite{Bao2024SimBase} for pretraining.
Let the model predict $U$ temporal boundaries $p_{ij}^{u}$ for the sentence query in the $i$-th data sample at the $j$-th epoch.
If the memory bank $\mathcal{M}_i$ contains $N_j$ instances at the $j$-th epoch, we compute the consensus score $c_r$ for the $r$-th memory instance $m_{r}$ by summing its Intersection over Union (IoU) with the other $N_j - 1$ instances in the memory bank as:
\begin{equation}
c_{r} = \sum_{k=1, k \neq r}^{N_{j}}\sigma(m_{r}, m_k)
\end{equation}
where $\sigma$ denotes the IoU operator.
Rather than directly using the temporal boundary $\hat{b}_i$ as pseudo ground truth, which is still prone to errors, we use the consensus scores $c_{r}$ to determine the most reliable pseudo ground truth from the memory bank.
The instance $m_{r^*}$ with the highest consensus is selected as the pseudo ground truth to correct $\hat{b}_i$ as:
\begin{equation}
r^* = \text{argmax}_r (c_{r})
\end{equation}
Next, we determine which prediction $p_{ij}^{u}$ to insert into the memory bank $\mathcal{M}_i$.
In detail, we use the confidence scores predicted by the model and select $u^*$, the one with the highest confidence score to insert into the memory bank $\mathcal{M}_i$:
\begin{equation}
u^* = \text{argmax}_u (\text{f}_{u}),
\end{equation}
where $\text{f}_u$ is the confidence score for the $u$-th prediction.

Finally, using the memory instance $m_{r^*}$ with consensus, the pretraining loss function is defined as:
{\small
\begin{equation}
\mathcal{L}_\text{pretrain} =  \lambda \sum_{u} \mathcal{L}_{\text{SimBase}} (p_{ij}^{u}, m_{r^*}) +  (1-\lambda) \sum_{u}\mathcal{L}_{\text{SimBase}} (p_{ij}^{u}, \hat{b}_i),
\end{equation}
}
where $\lambda$ is a hyperparameter to balance the loss terms
and $\mathcal{L}_{\text{SimBase}}$ is the loss function as defined in SimBase~\cite{Bao2024SimBase}.

\begin{table*}[t!]
\centering
\caption{
Performance comparison of state-of-the-art methods in \textit{zero-shot}, \textit{fully-supervised}, and \textit{unsupervised} settings, referred to as ZS, Full, and Unsup, respectively.
``Pretrain'' indicates whether the model is pretrained on video-language data.
The gray row represents ReCorrect's performance as a percentage of the previous best \textit{fully-supervised} method SimBase.
}
\label{table_zero_shot}
\scalebox{1.0}{
\begin{tabular}{clccccccccccccc}
\toprule
\multirow{2}{*}{Setting} &\multirow{2}{*}{Method}  &\multirow{2}{*}{Pretrain}
&\multicolumn{4}{c}{Charades STA} 
&\multicolumn{4}{c}{ActivityNet Captions}  
\\ \cmidrule(lr){4-7} \cmidrule(lr){8-11}
&&&R@0.3   &R@0.5   &R@0.7 &mIoU   &R@0.3  &R@0.5  &R@0.7 &mIoU\\
\midrule
\multirow{10}{*}{ZS} &       Luo et al.~\cite{Luo2023ZeroShotVM} & \xmark &          56.77 &          42.93 &          20.13 &          37.92 &          48.28 &          27.90 &          11.57 &          32.37 \\
&         Lu et al.~\cite{Lu2024ZeroShotVG} & \xmark &          47.74 &          34.62 &          20.16 &          32.97 &          49.26 &          31.45 &          15.27 &          33.25 \\
&     VideoChat-7B~\cite{Li2023VideoChatCV} & \cmark &           9.00 &           3.30 &           1.30 &           6.50 &           8.80 &           3.70 &           1.50 &           7.20 \\
&VideoLLaMA-7B~\cite{Zhang2023VideoLLaMAAI} & \cmark &          10.40 &           3.80 &           0.90 &           7.10 &           6.90 &           2.10 &           0.80 &           6.50 \\
&      VideoChatGPT-7B~\cite{maaz2023video} & \cmark &          20.00 &           7.70 &           1.70 &          13.70 &          26.40 &          13.60 &           6.10 &          18.90 \\
&             VTG-GPT~\cite{Xu2024VTGGPTTZ} & \cmark &          59.48 &          43.68 &          25.94 &          39.81 &          47.13 &          28.25 &          12.84 &          30.49 \\
&             UniVTG~\cite{Lin2023UniVTGTU} & \cmark &          44.09 &          25.22 &          10.03 &          27.12 &            $-$ &            $-$ &            $-$ &            $-$ \\
&                         GPT4o Pretraining & \cmark &          61.77 &          45.46 &          23.10 &          41.43 &          49.15 &          28.28 &          13.52 &          33.21 \\
&                 \textbf{ReCorrect (Ours)} & \cmark & \textbf{66.54} & \textbf{51.15} & \textbf{28.54} & \textbf{45.63} & \textbf{54.68} & \textbf{33.35} & \textbf{15.15} & \textbf{35.96} \\
\rowcolor{lightgray} & \multicolumn{2}{l}{Relative to SimBase} &         85.6\% &         76.9\% &         64.8\% &         81.3\% &         85.5\% &         67.6\% &         49.7\% &         76.4\% \\
\noalign{\vskip-2pt}
\midrule
\multirow{6}{*}{Full} &        UnLoc~\cite{Yan2023UnLocAU} & \xmark &            $-$ &          60.80 &          38.40 &            $-$ &            $-$ &          48.00 &          30.20 &            $-$ \\
&       MESM~\cite{Liu2023TowardsBA} & \xmark &            $-$ &          61.24 &          38.04 &            $-$ &            $-$ &            $-$ &            $-$ &            $-$ \\
&   BAM-DETR~\cite{Lee2023BAMDETRBM} & \xmark &          72.93 &          59.95 &          39.38 &          52.33 &            $-$ &            $-$ &            $-$ &            $-$ \\
&     SimBase~\cite{Bao2024SimBase}  & \xmark &          77.77 &          66.48 &          44.01 &          56.15 &          63.98 &          49.35 &          30.48 &          47.07 \\
&        SimBase + GPT4o Pretraining & \cmark & \textbf{78.79} &          68.20 &          44.09 &          56.96 &          64.72 &          49.18 &          30.67 &          47.42 \\
&\textbf{SimBase + ReCorrect (Ours)} & \cmark &          78.55 & \textbf{68.39} & \textbf{45.78} & \textbf{57.42} & \textbf{65.12} & \textbf{49.45} & \textbf{30.73} & \textbf{47.59} \\
\midrule
\multirow{12}{*}{Unsup} &     Gao et al~\cite{Gao2021LearningVM} & \xmark &          46.69 &          20.14 &           8.27 &            $-$ &          46.15 &          26.38 &          11.64 &            $-$ \\
&          PSVL~\cite{Nam2021ZeroshotNL} & \xmark &          46.47 &          31.29 &          14.17 &          31.24 &          44.74 &          30.08 &          14.74 &          29.62 \\
&     PZVMR~\cite{Wang2022PromptbasedZV} & \xmark &          46.83 &          33.21 &          18.51 &          32.62 &          45.73 &          31.26 &          17.84 &          30.35 \\
&Kim et al.~\cite{Kim2022LanguagefreeTF} & \xmark &          52.95 &          37.24 &          19.33 &          36.05 &          47.61 &          32.59 &          15.42 &          31.85 \\
&  CoroNet~\cite{Holla2023CommonsenseFZ} & \xmark &          49.21 &          34.60 &          17.93 &          32.73 &          46.05 &          28.19 &          12.84 &          31.11 \\
&       SPL~\cite{Zheng2023GeneratingSP} & \xmark &          60.73 &          40.70 &          19.62 &          40.47 &          50.24 &          27.24 &          15.03 &          35.44 \\
&                      GPT4o Finetuning  & \xmark &          61.24 &          44.51 &          22.11 &          40.91 &          49.33 &          28.94 &          13.20 &          33.10 \\
&                  ReCorrect Finetuning  & \xmark &          65.75 &          47.32 &          25.83 &          44.48 &          55.30 &          35.64 &          17.38 &          37.89 \\
&         ProTeGe~\cite{Wang2023ProTGUP} & \cmark &          46.79 &          31.84 &          17.51 &          31.25 &          45.02 &          27.85 &          14.89 &          33.04 \\
&         GPT4o Finetuning + Pretraining & \cmark &          65.72 &          49.10 &          25.21 &          44.22 &          50.58 &          30.56 &          14.13 &          34.09 \\
&              \textbf{ReCorrect (Ours)} & \cmark & \textbf{70.96} & \textbf{54.42} & \textbf{31.10} & \textbf{48.66} & \textbf{58.31} & \textbf{37.83} & \textbf{18.57} & \textbf{39.74} \\
\rowcolor{lightgray} & \multicolumn{2}{l}{Relative to SimBase} &         91.2\% &         81.9\% &         70.7\% &         86.7\% &         91.1\% &         76.7\% &         60.9\% &         84.4\% \\
\noalign{\vskip-2pt}
\bottomrule
\end{tabular}
}
\end{table*}

\subsection{Finetuning on Various Settings}
The pretrained ReCorrect model can be seamlessly adapted to various downstream settings on the target dataset for video moment retrieval, such as zero-shot inference, unsupervised learning, fully-supervised learning.

\noindent
\textbf{Zero-Shot Seting.}
The pretrained models are applied directly to the target datasets without fine-tuning, meaning the model operates without access to any  videos or annotations from the target dataset.

\noindent
\textbf{Unsupervised Setting.}
Only unlabeled videos from the target dataset, are used to finetune the pretrained models.
First, we generate pseudo annotations for these unlabeled videos following  Vid-Morp, and then finetune the pretrained model using ReCorrect algorithm.
The loss function for unsupervised finetuning is defined as:
\begin{equation}
\mathcal{L}_\text{unsup} = \mathcal{L}_\text{pretrain} (p^\text{unsup}, \hat{b}^\text{unsup})
\end{equation}
where $p^\text{unsup}$ denotes the model's prediction, and $\hat{b}^\text{unsup}$ represents the pseudo temporal boundary, which is progressively refined by ReCorrect.

\noindent
\textbf{Fully-Supervised Setting.}
We finetune the pretrained ReCorrect model on the target dataset using full manual annotations.
The loss function is the same as that used in the fully-supervised method SimBase~\cite{Bao2024SimBase}:
\begin{equation}
\mathcal{L}_\text{full} = \mathcal{L}_\text{SimBase} (p^\text{full}, b^\text{full})
\end{equation}
where $p^\text{full}$ represents the model's prediction, and $b^\text{full}$ corresponds to the manual annotation.

\section{Experiment}
\subsection{Datasets and Evaluation Metrics}
We evaluate the performance of the proposed methods on two large-scale datasets: Charades-STA~\cite{tall} and ActivityNet Captions~\cite{dense_cap}.
We adopt the evaluation metric `R@m' for video moment retrieval to evaluate the performance.
Specifically, we calculate the Intersection over Union (IoU) between the retrieved temporal moment and the ground truth.
Then `R@m' is defined as the percentage of language queries having correct moment retrieval results, where a moment retrieval is correct if its IoU is larger than $m$.

\subsection{Implementation Details}
We use the pretrained CLIP~\cite{Radford2021LearningTV} model to extract visual and textual features.
The network architecture of the pretrained model is the same as that of SimBase~\cite{Bao2024SimBase}.
The hyperparameter cleaning ratio $R$ is set to 40\%.
The frame number $T$ and step size $\delta$ for the semantics-guided refinement are set to $256$ and $5$, respectively.
The hyperparameters $\alpha_1$ and $\alpha_2$ are configured as $0.22$ and $0.92$, respectively.
We train our model using the Adam optimizer~\cite{kingma2014adam} with a batch size of $256$ and a learning rate of $0.0004$.
The pretraining epoch number is set to $15$.
The loss weight $\lambda$ is set to $0.7$.
For the video moment retrieval model, we adopt the same network architecture as the state-of-the-art fully-supervised model SimBase~\cite{Bao2024SimBase}.
Details of the network architecture can be referred to~\cite{Bao2024SimBase} and can also be accessed through our implementation at \href{https://github.com/baopj/Vid-Morp}{https://github.com/baopj/Vid-Morp}.

\subsection{Performance Comparisons}
\subsubsection{Zero-Shot Inference}
%
Previous zero-shot methods for video moment retrieval (VMR) can be categorized into three main classes:
i) Adapting CLIP models pretrained on image-text corpora to VMR tasks, as seen in~\cite{Luo2023ZeroShotVM, Lu2024ZeroShotVG}.
ii) Large video-language models such as~\cite{Li2023VideoChatCV, Zhang2023VideoLLaMAAI, maaz2023video}.
iii) Ensembles of multimodal language models for VMR, including VTG-GPT~\cite{Xu2024VTGGPTTZ}.

As illustrated in the first part of Table~\ref{table_zero_shot}, our ReCorrect method outperforms all previous zero-shot approaches by a clear margin.
For instance, it surpasses the previous best model, VTG-GPT, by 7 points in R@0.7 and over 5 points in mIoU on both datasets.
When directly pretraning on the original pseudo labels of GPT4o on Vid-Morp dataset, GPT4o Pretraining achieves similar results to VTG-GPT and Lu et al.~\cite{Lu2024ZeroShotVG} on both datasets.
And ReCorrect consistently boosts GPT4o Pretraining's performance across both datasets, highlighting the critical role of addressing various types of errors in pseudo labels.

\subsubsection{Fully-Supervised Learning}
%
The second part of Table~\ref{table_zero_shot} shows that adding both GPT4o Pretraining and ReCorrect to SimBase~\cite{Bao2024SimBase} in a fully-supervised setting enhance performance.
Both methods show an R@0.5 improvement of over 1.5 points on Charades and around 1 point on ActivityNet for R@0.3 compared to SimBase.
Compared to GPT4o Pretraining, ReCorrect further boosts performance by approximately 2 points at R@0.7 on Charades and 0.5 points at R@0.3 on ActivityNet.
The margin of improvement between ReCorrect and GPT4o Pretraining is narrower here than in zero-shot settings, as the fine-tuning datasets provide 12.8K and 37.4K manual labels, respectively.

\subsubsection{Unsupervised Learning}
%
The third part of Table~\ref{table_zero_shot} summarizes the performance comparisons in unsupervised learning settings.
The methods ``GPT4o Finetuning'' and ``ReCorrect Finetuning'' represent models that are not pretrained on Vid-Morp datasets but instead fine-tuned solely on Charades or ActivityNet.
Their pretrained versions demonstrate approximately 3 and 5 points improvements over the non-pretrained versions.
It is notable that the unsupervised ReCorrect method achieves approximately 85\% of the overall performance of the fully-supervised SimBase on both datasets in terms of metrics mIoU.
Such a close margin between unsupervised and fully-supervised methods underscores the potential of our Vid-Morp dataset to relieve the manual annotation requirement in the VMR task.

\subsubsection{Comparisons to Existing Pretraining Paradigms}
%
Here we compare ReCorrect with existing pretraining paradigms, \textit{i.e.,}  UniTVG~\cite{Lin2023UniVTGTU} and ProTeGe~\cite{Wang2023ProTGUP}.

\noindent
\textit{UniTVG}~\cite{Lin2023UniVTGTU} presents a general-purpose pretraining approach  unified for a list of video-language temporal understanding tasks, including video moment retrieval.
However, as shown in the first part of Table~\ref{table_zero_shot}, its zero-shot performance is significantly lower than that of most apporaches specific to VMR, such as Luo et al.~\cite{Luo2023ZeroShotVM}, as well as our ReCorrect.
For example, UniTVG achieves only half of the R@0.7 score of Luo et al.~\cite{Luo2023ZeroShotVM} and one-third of ReCorrect, underscoring the advantage of designing pretraining datasets specifically tailored to VMR.

\noindent
\textit{ProTeGe}~\cite{Wang2023ProTGUP} introduces a pretraining paradigm  on the feature extraction backbone for video moment retrieval.
In contrast, ReCorrect focuses on pretraining the VMR model itself while using a fixed backbone.
This enables our model to support zero-shot inference, which ProTeGe does not.
In unsupervised settings, as presented in the second part of Table~\ref{table_zero_shot}, ReCorrect demonstrates a substantial advantage over ProTeGe.
For instance, ReCorrect achieves 78\% in R@0.7 on Charades, highlighting the importance of pretraining the VMR model itself instead of soly the feature extraction backbone.

\begin{table*}[t!]
\centering
\caption{
Performance comparisons on two types of out-of-distribution datasets~\cite{Li2022CompositionalTG}: Novel Composition and Novel Words.
}
\label{table_novel_charades}
\scalebox{0.95}{
\begin{tabular}{cccccccccccccc}
\toprule
\multirow{3}{*}{Method} &\multirow{3}{*}{Setting}
&\multicolumn{6}{c}{Charades-CG} 
&\multicolumn{6}{c}{ActivityNet-CG} 
\\ \cmidrule(lr){3-8} \cmidrule(lr){9-14}
& 
&\multicolumn{3}{c}{Novel Composition} 
&\multicolumn{3}{c}{Novel Word}  
&\multicolumn{3}{c}{Novel Composition} 
&\multicolumn{3}{c}{Novel Word} 
\\ \cmidrule(lr){3-5} \cmidrule(lr){6-8} \cmidrule(lr){9-11} \cmidrule(lr){12-14}
& & R@0.5   &R@0.7 &mIoU     &R@0.5  &R@0.7 &mIoU & R@0.5   &R@0.7 &mIoU     &R@0.5  &R@0.7 &mIoU\\ 
\midrule
WSSL~\cite{Duan2018WeaklySD} & \multirow{1}{*}{Weak} &3.61 & 1.21 & 8.26 & 2.79 & 0.73 & 7.92 & 2.89 & 0.76 & 7.65 & 3.09 & 1.13 & 7.10 \\
\midrule
TSP-PRL~\cite{Wu2020TreeStructuredPB} & \multirow{9}{*}{Full} &16.30 & 2.04 & 13.52 & 14.83 & 2.61 & 14.03 & 14.74 & 1.43 & 12.61 & 18.05 & 3.15 & 14.34 \\
TMN~\cite{Liu2018TemporalMN} & &8.68 & 4.07 & 10.14 & 9.43 & 4.96 & 11.23 & 8.74 & 4.39 & 10.08 & 9.93 & 5.12 & 11.38 \\
2D-TAN~\cite{Zhang2020Learning2T} & &30.91 & 12.23 & 29.75 & 29.36 & 13.21 & 28.47 & 22.80 & 9.95 & 28.49 & 23.86 & 10.37 & 28.88 \\
LGI~\cite{Mun2020LocalGlobalVI} & &29.42 & 12.73 & 30.09 & 26.48 & 12.47 & 27.62 & 23.21 & 9.02 & 27.86 & 23.10 & 9.03 & 26.95 \\
VLSNet~\cite{Zhang2020SpanbasedLN} & &24.25 & 11.54 & 31.43 & 25.60 & 10.07 & 30.21 & 20.21 & 9.18 & 29.07 & 21.68 & 9.94 & 29.58 \\
DeCo~\cite{Yang2023DeCoDA} & &47.39 & 21.06 & 40.70 & $-$ & $-$ & $-$ & 28.69 & 12.98 & 32.67 & $-$ & $-$ & $-$ \\
VISA~\cite{Li2022CompositionalTG} & &45.41 & 22.71 & 42.03 & 42.35 & 20.88 & 40.18 & \textbf{31.51} & \textbf{16.73} & \textbf{35.85} & \textbf{30.14} & \textbf{15.90} & \textbf{35.13} \\
2D-TAN+SSL~\cite{Li2023ExploringTE} & &35.42 & 17.95 & 33.07 & 43.60 & 25.32 & 39.32 & $-$ & $-$ & $-$ & $-$ & $-$ & $-$ \\
MS-2D-TAN+SSL~\cite{Li2023ExploringTE} & &\textbf{46.54} & \textbf{25.10} & \textbf{40.00} & \textbf{50.36} & \textbf{28.78} & \textbf{43.15} & $-$ & $-$ & $-$ & $-$ & $-$ & $-$ \\
\midrule
Luo et al.~\cite{Luo2023ZeroShotVM} & \multirow{3}{*}{ZS} &$-$ & $-$ & $-$ & 45.04 & 21.44 & $-$ & $-$ & $-$ & $-$ & 24.57 & 10.54 & $-$ \\
GPT4o Pretraining & &40.35 & 18.94 & 38.40 & 48.06 & 25.18 & 43.15 & 24.67 & 10.38 & 29.34 & 24.44 & 10.18 & 29.41 \\
\textbf{ReCorrect (Ours)} & &\textbf{48.20} & \textbf{25.10} & \textbf{43.79} & \textbf{53.96} & \textbf{29.06} & \textbf{46.67} & \textbf{29.90} & \textbf{13.19} & \textbf{32.76} & \textbf{30.36} & \textbf{12.81} & \textbf{32.63} \\
\bottomrule
\end{tabular}
}
\end{table*}
\begin{table*}[t!]
\centering
\caption{
Performance comparisons on datasets with Changing Distribution (CD)~\cite{Yuan2021ACL} of temporal boundaries.
}
\label{table_ood}
\scalebox{1.0}{
\begin{tabular}{cccccccccccccc}
\toprule
\multirow{2}{*}{Method} &\multirow{2}{*}{Setting}
&\multicolumn{3}{c}{Charades CD} 
&\multicolumn{3}{c}{ActivityNet CD}  
\\ \cmidrule(lr){3-5} \cmidrule(lr){6-8}
& & R@0.3   &R@0.5 &R@0.7     & R@0.3   &R@0.5 &R@0.7\\ 
\midrule
WSSL~\cite{Duan2018WeaklySD} & \multirow{1}{*}{Weak} &35.86 & 23.67 & 8.27 & 17.00 & 7.17 & 1.82 \\
\midrule
TSP-PRL~\cite{Wu2020TreeStructuredPB} & \multirow{5}{*}{Full} &31.93 & 19.37 & 6.20 & 29.61 & 16.63 & 7.43 \\
ABLR~\cite{Yuan2018ToFW} & &44.62 & 31.57 & 11.38 & 33.45 & 20.88 & 10.03 \\
2D-TAN~\cite{Zhang2020Learning2T} & &43.45 & 30.77 & 11.75 & 30.86 & 18.38 & 9.11 \\
DRN~\cite{Zeng2020DenseRN} & &40.45 & 30.43 & 15.91 & 36.86 & \textbf{25.15} & \textbf{14.33} \\
MomentDETR~\cite{Lei2021QVHighlightsDM} & &\textbf{57.34} & \textbf{41.18} & \textbf{19.31} & \textbf{39.98} & 21.30 & 10.58 \\
\midrule
GPT4o Pretraining & \multirow{2}{*}{ZS} &60.68 & 40.84 & 14.12 & 40.57 & 23.45 & 9.81 \\
\textbf{ReCorrect (Ours)} & &\textbf{65.98} & \textbf{46.80} & \textbf{21.38} & \textbf{45.37} & \textbf{26.68} & \textbf{12.06} \\
\bottomrule
\end{tabular}
}
\end{table*}

\subsubsection{Out-of-Distribution Scenarios}
%
Table~\ref{table_novel_charades} and~\ref{table_ood} present performance comparisons across three types of out-of-distribution datasets~\cite{Li2022CompositionalTG,Yuan2021ACL}: Novel Composition, Novel Word, and Changing Distribution of temporal boundary.
Our ReCorrect method notably outperforms SSL~\cite{Li2023ExploringTE} and VISA~\cite{Li2022CompositionalTG} on Novel Composition and Novel Word datasets.
For the Changing Distribution dataset, specifically Charades-CD, ReCorrect also exceeds MomentDETR~\cite{Lei2021QVHighlightsDM}.
On ActivityNet CD, ReCorrect performs comparably to MomentDETR, showing a 5-point advantage in R@0.3 but a 2-point disadvantage in R@0.7.

It's worth noting that methods like DeCo, VISA, and SSL are (1) fully supervised and (2) specifically tailored to adapt to out-of-distribution scenarios through carefully designed algorithms.
In contrast, zero-shot ReCorrect require no fine-tuning and does not have specific algorithmic designs for these scenarios.
However, our ReCorrect in a zero-shot manner, without any finetuning or specific algorithm level designing to these out-of-distribution scenarios.
This strong performance can be attributed to the scale and versatility of our Vid-Morp dataset, which encompasses a broad range of video content, annotations, and vocabulary (see Table~\ref{table_data_compare}), allowing ReCorrect to handle diverse distribution scenarios effectively.

\begin{figure}[t!]
\centering
\includegraphics[width=0.88\linewidth]{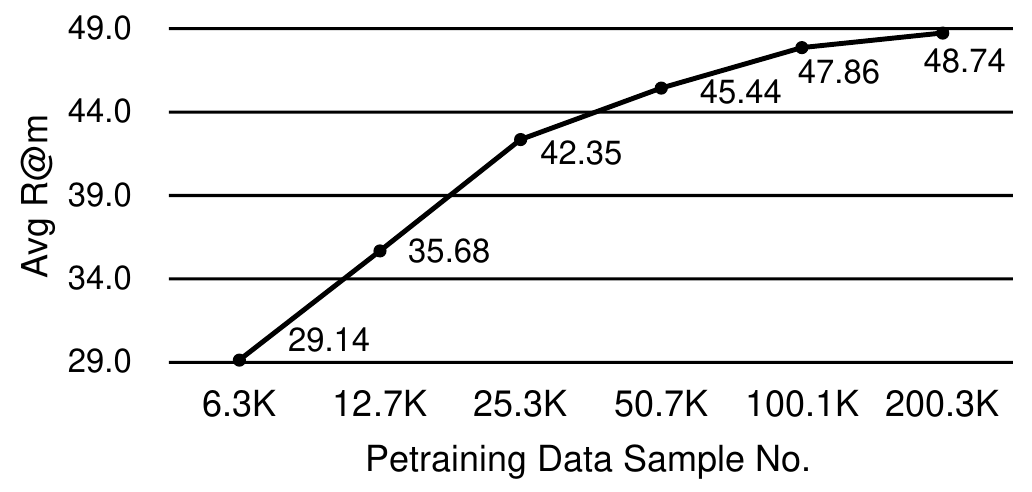}
\caption{
Scability of pretraining dataset size.
}
\label{fig_data_num}
\end{figure}

\subsubsection{Qualitative Comparisons}
\begin{figure*}[t!]
\centering
\includegraphics[width=\linewidth]{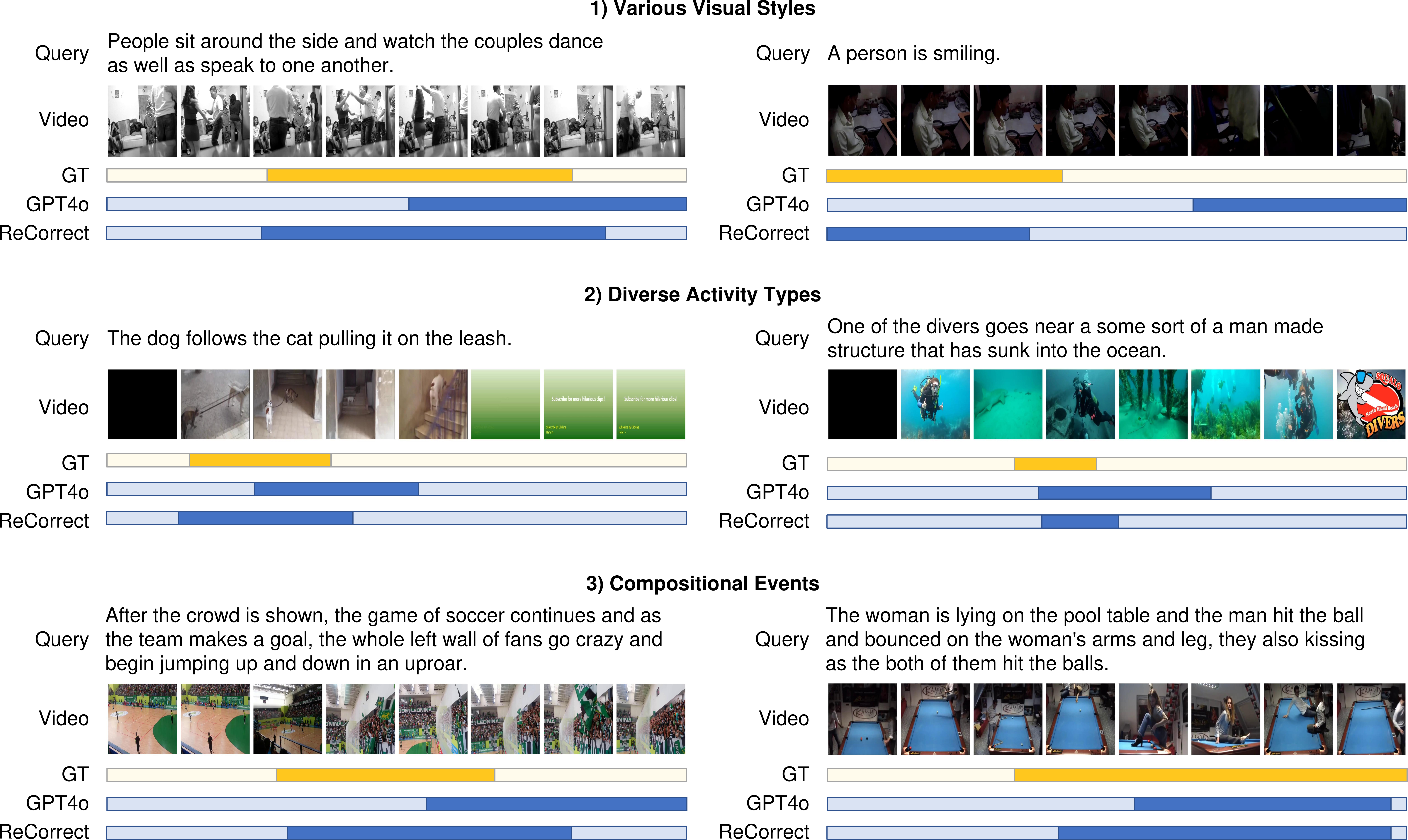}
\caption{
Qualitative comparison of zero-shot inference between GPT4o pretraining and our ReCorrect algorithm.
Our zero-shot ReCorrect demonstrates its powerful capability in video moment retrieval across:
1) various visual conditions, such as black-and-white movie segments and low-light scenarios;
2) diverse activity types, including animal behavior and underwater scenes; and
3) compositional events comprising multiple sub-events and necessitating temporal reasoning.
Here ``GT'' indicates ground truth.
The darker yellow rectangle represents the ground-truth temporal boundaries, while the darker blue rectangle denotes the model's prediction.
}
\label{fig_qualitive}
\end{figure*}

Fig.~\ref{fig_qualitive} illustrates a qualitative performance comparison between GPT-4o pretraining and our ReCorrect algorithm in zero-shot video moment retrieval.
The results highlight the strengths of our zero-shot ReCorrect approach across three challenging scenarios: 
(1) handling diverse visual conditions, such as black-and-white movie segments and low-light scenarios; 
(2) effectively retrieving moments from diverse activity types, including animal behavior and underwater scenes; 
and 
(3) accurately reasoning about compositional events that involve multiple sub-events and require temporal understanding. 

\begin{figure}[t!]
\centering
\includegraphics[width=0.88\linewidth]{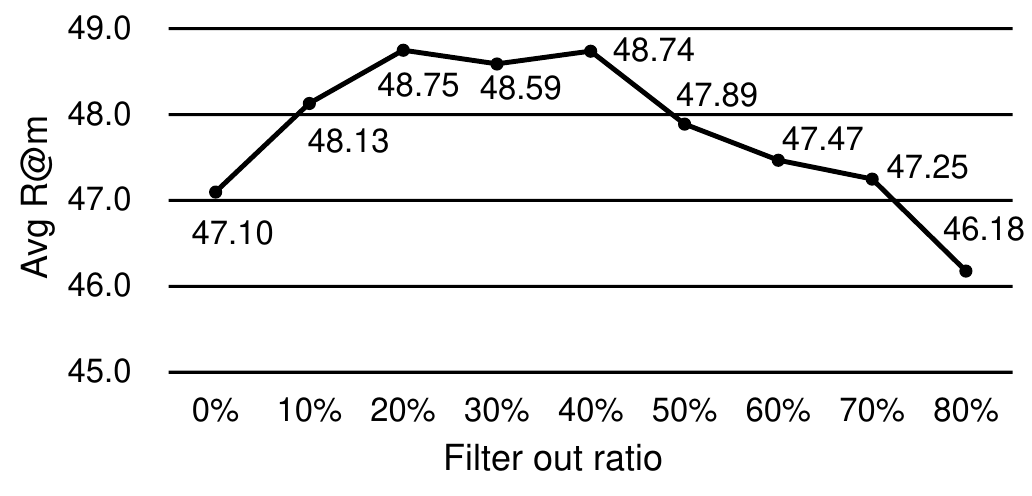}
\caption{
Ablation studies on the cleaning ratio.
}
\label{fig_filter_out}
\end{figure}

\subsection{Ablation Studies}
To assess the effectiveness of the proposed ReCorrect algorithm, we conduct ablation studies on the Charades-STA.

\subsubsection{Scability of Pretraining Dataset Size}
%
Fig.~\ref{fig_data_num} illustrates zero-shot performance against the number of data samples for pretraining on a semi-logarithmic scale.
We evaluate the overall performance as the average of R@m values, where $m={0.3, 0.5, 0.7}$.
A linear increase in performance is observed as the pretraining data size doubles, starting from 6.3K to 12.7K. This trend continues as the data size grows from 25.3K to 200.3K, although the slope of the increase becomes less steep at higher scales.
These results demonstrate that our Vid-Morp dataset exhibits scalable performance improvements.

\subsubsection{Impact of Cleaning Ratio}
%
As Vid-Morp dataset is collected with minimal human intervention, it inevitably includes errors such as idle videos and mismatched video-query pairs, as shown in Fig.~\ref{fig_benchmark_challenge}.
To address this, ReCorrect algorithm incorporates a label-cleaning module in the semantics-guided refinement phase, where the cleaning ratio determines the percentage of data samples filtered out.
Fig.~\ref{fig_filter_out} illustrates the impact of the cleaning ratio on zero-shot performance.
Performance enhances as the cleaning ratio increases from 0\% to 30\%, as samples with pseudo-label errors are removed.
However, performance declines when the cleaning ratio exceeds 50\%.
The curve also indicates that a cleaning ratio between 20\% and 40\% yields satisfactory results.

\begin{table}[t!]
\centering
\caption{
Ablation studies on zero-shot inference.
}
\label{table_ablation_modules_zs}
\scalebox{1.0}{
\begin{tabular}{ccccccccccccccc}
\toprule
Clean & Adjust & Correct  & R@0.3  & R@0.5  & R@0.7 & mIoU\\ 
\midrule
\xmark & \xmark & \xmark &          61.77 &          45.46 &          23.10 &          41.43 \\
\cmark & \xmark & \xmark &          64.96 &          48.00 &          23.86 &          42.94 \\
\xmark & \cmark & \xmark &          65.27 &          48.28 &          25.63 &          44.17 \\
\cmark & \cmark & \xmark &          65.83 &          49.46 &          26.82 &          44.45 \\
\cmark & \cmark & \cmark & \textbf{66.54} & \textbf{51.15} & \textbf{28.54} & \textbf{45.63} \\\bottomrule
\end{tabular}
}
\end{table}

\subsubsection{Effectiveness of the Proposed Modules}
%
Our ReCorrect algorithm comprises three major modules: 1) label cleaning, 2) boundary adjustment in the semantics-guided refinement phase, and 3) memory consensus correction.
These modules are utilized for both pretraining and unsupervised learning.
To evaluate their effectiveness, we investigate their impact on both zero-shot inference and unsupervised learning.

Table~\ref{table_ablation_modules_zs} demonstrates the effectiveness of each module for zero-shot inference performance, where the three marjor modules are denoted as “Clean,” “Adjust,” and “Correct” respectively.
Table~\ref{table_ablation_modules_unsup} further studies the effectiveness of these modules on unsupervised fine-tuning.
These results consistently show that each module contributes positively to performance in both pretraining and unsupervised learning, and removing any one of them leads to a notable decline in performance.

\begin{table}[t!]
\centering
\caption{
Ablation studies on unsupervised learning.
}
\label{table_ablation_modules_unsup}
\scalebox{0.9}{
\begin{tabular}{ccccccccccccccc}
\toprule
\multirow{2}{*}{Pretrain} &\multicolumn{3}{c}{Finetune} & \multirow{2}{*}{R@0.3}  & \multirow{2}{*}{R@0.7} & \multirow{2}{*}{R@0.7} & \multirow{2}{*}{mIoU}\\
\cmidrule(lr){2-4}
&Clean & Adjust & Correct  &&&\\
\midrule
\xmark & \xmark & \xmark & \xmark &          61.24 &          44.51 &          22.11 &          40.91 \\
\cmark & \xmark & \xmark & \xmark &          67.52 &          50.70 &          26.28 &          45.37 \\
\cmark & \cmark & \xmark & \xmark &          68.62 &          51.89 &          27.27 &          46.35 \\
\cmark & \cmark & \cmark & \xmark &          69.52 &          53.21 &          30.17 &          47.61 \\
\cmark & \cmark & \cmark & \cmark & \textbf{70.96} & \textbf{54.42} & \textbf{31.10} & \textbf{48.66} \\\bottomrule
\end{tabular}
}
\end{table}

\section{Conclusion}
%
This paper introduces Vid-Morp, a large-scale dataset for Video Moment Retrieval Pretraining, collected with minimal manual intervention.
To address three types of errors in Vid-Morp's pseudo labels, we propose the Refinement and Correction (ReCorrect) algorithm.
It consists of 1) semantics-guided refinement, which filters unpaired data and adjusts temporal boundaries, and 2) memory-consensus correction, where a memory bank tracks predictions to correct boundaries based on consensus.
Our experiments demonstrate the effectiveness of ReCorrect in various learning settings, showing its strong generalizability.

\bibliography{meta/citations.bib}

\begin{thebibliography}{10}
\providecommand{\url}[1]{#1}
\csname url@samestyle\endcsname
\providecommand{\newblock}{\relax}
\providecommand{\bibinfo}[2]{#2}
\providecommand{\BIBentrySTDinterwordspacing}{\spaceskip=0pt\relax}
\providecommand{\BIBentryALTinterwordstretchfactor}{4}
\providecommand{\BIBentryALTinterwordspacing}{\spaceskip=\fontdimen2\font plus
\BIBentryALTinterwordstretchfactor\fontdimen3\font minus
  \fontdimen4\font\relax}
\providecommand{\BIBforeignlanguage}[2]{{%
\expandafter\ifx\csname l@#1\endcsname\relax
\typeout{** WARNING: IEEEtran.bst: No hyphenation pattern has been}%
\typeout{** loaded for the language `#1'. Using the pattern for}%
\typeout{** the default language instead.}%
\else
\language=\csname l@#1\endcsname
\fi
#2}}
\providecommand{\BIBdecl}{\relax}
\BIBdecl

\bibitem{tall}
J.~Gao, C.~Sun, Z.~Yang, and R.~Nevatia, ``Tall: Temporal activity localization
  via language query,'' in \emph{ICCV}, 2017.

\bibitem{dense_cap}
R.~Krishna, K.~Hata, F.~Ren, L.~Fei-Fei, and J.~Carlos~Niebles,
  ``Dense-captioning events in videos,'' in \emph{ICCV}, 2017.

\bibitem{Qi2021SemanticsAwareSB}
M.~Qi, J.~Qin, Y.~Yang, Y.~Wang, and J.~Luo, ``Semantics-aware spatial-temporal
  binaries for cross-modal video retrieval,'' \emph{TIP}, 2021.

\bibitem{Sreenu2019IntelligentVS}
G.~Sreenu and M.~A.~S. Durai, ``Intelligent video surveillance: a review
  through deep learning techniques for crowd analysis,'' \emph{Journal of Big
  Data}, 2019.

\bibitem{Jin2024Weak}
Y.~Jin and Y.~Mu, ``Weakly-supervised spatio-temporal video grounding with
  variational cross-modal alignment,'' in \emph{ECCV}, 2024.

\bibitem{Zhu2021DSNetAF}
W.~Zhu, J.~Lu, J.~Li, and J.~Zhou, ``Dsnet: A flexible detect-to-summarize
  network for video summarization,'' \emph{TIP}, 2021.

\bibitem{attention_2}
M.~Liu, X.~Wang, L.~Nie, Q.~Tian, B.~Chen, and T.-S. Chua, ``Cross-modal moment
  localization in videos,'' in \emph{ACM MM}, 2018.

\bibitem{Mun2020LocalGlobalVI}
J.~Mun, M.~Cho, and B.~Han, ``Local-global video-text interactions for temporal
  grounding,'' in \emph{CVPR}, 2020.

\bibitem{cbp}
J.~Wang, L.~Ma, and W.~Jiang, ``Temporally grounding language queries in videos
  by contextual boundary-aware prediction,'' in \emph{AAAI}, 2020.

\bibitem{man}
D.~Zhang, X.~Dai, X.~Wang, Y.-F. Wang, and L.~S. Davis, ``Man: Moment alignment
  network for natural language moment retrieval via iterative graph
  adjustment,'' in \emph{CVPR}, 2019.

\bibitem{Cai2024TemporalSG}
C.~Cai, R.~Zhang, J.~Gao, K.~Wu, K.-H. Yap, and Y.~Wang, ``Temporal sentence
  grounding with temporally global textual knowledge,'' in \emph{ICME}, 2024.

\bibitem{Bao2024SimBase}
P.~Bao and A.~Kot, ``Simbase: A simple baseline for temporal video grounding,''
  \emph{arXiv}, 2024.

\bibitem{Zhang2020Learning2T}
S.~Zhang, H.~Peng, J.~Fu, and J.~Luo, ``Learning 2d temporal adjacent networks
  for moment localization with natural language,'' in \emph{AAAI}, 2020.

\bibitem{Zhang2019CrossModalIN}
Z.~Zhang, Z.~Lin, Z.~Zhao, and Z.~Xiao, ``Cross-modal interaction networks for
  query-based moment retrieval in videos,'' 2019.

\bibitem{Bao2022LearningSI}
P.~Bao and Y.~Mu, ``Learning sample importance for cross-scenario video
  temporal grounding,'' 2022.

\bibitem{Bao2021DenseEG}
P.~Bao, Q.~Zheng, and Y.~Mu, ``Dense events grounding in video,'' in
  \emph{AAAI}, 2021.

\bibitem{bao2024omnipotent}
P.~Bao, Z.~Shao, W.~Yang, B.~P. Ng, M.~H. Er, and A.~C. Kot, ``Omnipotent
  distillation with llms for weakly-supervised natural language video
  localization: When divergence meets consistency,'' in \emph{AAAI}, 2024.

\bibitem{bao2024local}
P.~Bao, Y.~Xia, W.~Yang, B.~P. Ng, M.~H. Er, and A.~C. Kot, ``Local-global
  multi-modal distillation for weakly-supervised temporal video grounding,'' in
  \emph{AAAI}, 2024.

\bibitem{Bao2023CrossModalLC}
P.~Bao, W.~Yang, B.~P. Ng, M.~H. Er, and A.~C. Kot, ``Cross-modal label
  contrastive learning for unsupervised audio-visual event localization,'' in
  \emph{AAAI}, 2023.

\bibitem{Yuan2021ACL}
Y.~Yuan, X.~Lan, L.~Chen, W.~Liu, X.~Wang, and W.~Zhu, ``A closer look at
  temporal sentence grounding in videos: Dataset and metric,'' in \emph{ACM MM
  Workshop}, 2021.

\bibitem{Li2022CompositionalTG}
J.~Li, J.~Xie, L.~Qian, L.~Zhu, S.~Tang, F.~Wu, Y.~Yang, Y.~Zhuang, and X.~E.
  Wang, ``Compositional temporal grounding with structured variational
  cross-graph correspondence learning,'' in \emph{CVPR}, 2022.

\bibitem{Nam2021ZeroshotNL}
J.~Nam, D.~Ahn, D.~Kang, S.~J. Ha, and J.~Choi, ``Zero-shot natural language
  video localization,'' in \emph{ICCV}, 2021.

\bibitem{Zheng2023GeneratingSP}
M.~Zheng, S.~Gong, H.~Jin, Y.~Peng, and Y.~Liu, ``Generating structured pseudo
  labels for noise-resistant zero-shot video sentence localization,'' in
  \emph{ACL}, 2023.

\bibitem{Kim2022LanguagefreeTF}
D.~Kim, J.~Park, J.~Lee, S.~H. Park, and K.~Sohn, ``Language-free training for
  zero-shot video grounding,'' in \emph{WACV}, 2022.

\bibitem{hendricks17iccv}
L.~A. Hendricks, O.~Wang, E.~Shechtman, J.~Sivic, T.~Darrell, and B.~Russell,
  ``Localizing moments in video with natural language,'' in \emph{ICCV}, 2017.

\bibitem{tacos}
M.~Regneri, M.~Rohrbach, D.~Wetzel, S.~Thater, B.~Schiele, and M.~Pinkal,
  ``Grounding action descriptions in videos,'' \emph{TACL}, 2013.

\bibitem{MM20}
S.~Zhang, J.~Su, and J.~Luo, ``Exploiting temporal relationships in video
  moment localization with natural language,'' in \emph{ACM MM}, 2019.

\bibitem{comp}
J.~C. Stroud, R.~McCaffrey, R.~Mihalcea, J.~Deng, and O.~Russakovsky,
  ``Compositional temporal visual grounding of natural language event
  descriptions.''

\bibitem{bao2024e3m}
P.~Bao, Z.~Shao, W.~Yang, B.~P. Ng, and A.~C. Kot, ``E3m: Zero-shot
  spatio-temporal video grounding with expectation-maximization multimodal
  modulation.''\hskip 1em plus 0.5em minus 0.4em\relax ECCV, 2024.

\bibitem{gong2020learning}
G.~Gong, X.~Wang, Y.~Mu, and Q.~Tian, ``Learning temporal co-attention models
  for unsupervised video action localization,'' in \emph{CVPR}, 2020.

\bibitem{soomro2017unsupervised}
K.~Soomro and M.~Shah, ``Unsupervised action discovery and localization in
  videos,'' in \emph{CVPR}, 2017.

\bibitem{Liao2024VideoINSTAZL}
R.~Liao, M.~Erler, H.~Wang, G.~Zhai, G.~Zhang, Y.~Ma, and V.~Tresp,
  ``Videoinsta: Zero-shot long video understanding via informative
  spatial-temporal reasoning with llms,'' in \emph{EMNLP}, 2024.

\bibitem{Wang2023ProTGUP}
L.~Wang, G.~Mittal, S.~Sajeev, Y.~Yu, M.~Hall, V.~N. Boddeti, and M.~Chen,
  ``Protege: Untrimmed pretraining for video temporal grounding by video
  temporal grounding,'' in \emph{CVPR}, 2023.

\bibitem{Lin2023UniVTGTU}
K.~Lin, P.~Zhang, J.~Chen, S.~Pramanick, D.~Gao, A.~Wang, R.~Yan, and M.~Z.
  Shou, ``Univtg: Towards unified video-language temporal grounding,'' in
  \emph{ICCV}, 2023.

\bibitem{Radford2021LearningTV}
A.~Radford, J.~W. Kim, C.~Hallacy, and et~al., ``Learning transferable visual
  models from natural language supervision,'' in \emph{ICML}, 2021.

\bibitem{Luo2023ZeroShotVM}
D.~Luo, J.~Huang, S.~Gong, H.~Jin, and Y.~Liu, ``Zero-shot video moment
  retrieval from frozen vision-language models,'' \emph{WACV}, 2023.

\bibitem{Lu2024ZeroShotVG}
Y.~Lu, R.~Quan, L.~Zhu, and Y.~Yang, ``Zero-shot video grounding with pseudo
  query lookup and verification,'' \emph{TIP}, 2024.

\bibitem{Li2023VideoChatCV}
K.~Li, Y.~He, Y.~Wang, Y.~Li, W.~Wang, P.~Luo, Y.~Wang, L.~Wang, and Y.~Qiao,
  ``Videochat: Chat-centric video understanding,'' \emph{ArXiv}, 2023.

\bibitem{Zhang2023VideoLLaMAAI}
H.~Zhang, X.~Li, and L.~Bing, ``Video-llama: An instruction-tuned audio-visual
  language model for video understanding,'' in \emph{EMNLP}, 2023.

\bibitem{maaz2023video}
M.~Maaz, H.~Rasheed, S.~Khan, and F.~S. Khan, ``Video-chatgpt: Towards detailed
  video understanding via large vision and language models.''

\bibitem{Xu2024VTGGPTTZ}
Y.~Xu, Y.~Sun, Z.~Xie, B.~Zhai, and S.~Du, ``Vtg-gpt: Tuning-free zero-shot
  video temporal grounding with gpt,'' \emph{Applied Sciences}, 2024.

\bibitem{Yan2023UnLocAU}
S.~Yan, X.~Xiong, A.~Nagrani, A.~Arnab, Z.~Wang, W.~Ge, D.~A. Ross, and
  C.~Schmid, ``Unloc: A unified framework for video localization tasks,'' in
  \emph{ICCV}, 2023.

\bibitem{Liu2023TowardsBA}
Z.~Liu, J.~Li, H.~Xie, P.~Li, J.~Ge, S.-A. Liu, and G.~Jin, ``Towards balanced
  alignment: Modal-enhanced semantic modeling for video moment retrieval,'' in
  \emph{AAAI}, 2024.

\bibitem{Lee2023BAMDETRBM}
P.~Lee and H.~Byun, ``Bam-detr: Boundary-aligned moment detection transformer
  for temporal sentence grounding in videos,'' in \emph{ECCV}, 2024.

\bibitem{Gao2021LearningVM}
J.~Gao and C.~Xu, ``Learning video moment retrieval without a single annotated
  video,'' \emph{TCSVT}, 2021.

\bibitem{Wang2022PromptbasedZV}
G.~Wang, X.~Wu, Z.~Liu, and J.~Yan, ``Prompt-based zero-shot video moment
  retrieval,'' in \emph{ACM MM}, 2022.

\bibitem{Holla2023CommonsenseFZ}
M.~Holla and I.~Lourentzou, ``Commonsense for zero-shot natural language video
  localization,'' in \emph{AAAI}, 2024.

\bibitem{Duan2018WeaklySD}
X.~Duan, W.~bing Huang, C.~Gan, J.~Wang, W.~Zhu, and J.~Huang, ``Weakly
  supervised dense event captioning in videos.''

\bibitem{Wu2020TreeStructuredPB}
J.~Wu, G.~Li, S.~Liu, and L.~Lin, ``Tree-structured policy based progressive
  reinforcement learning for temporally language grounding in video,'' in
  \emph{AAAI}, 2020.

\bibitem{Liu2018TemporalMN}
B.~Liu, S.~Yeung, E.~Chou, D.-A. Huang, L.~Fei-Fei, and J.~C. Niebles,
  ``Temporal modular networks for retrieving complex compositional activities
  in videos,'' in \emph{ECCV}, 2018.

\bibitem{Zhang2020SpanbasedLN}
H.~Zhang, A.~Sun, W.~Jing, and J.~T. Zhou, ``Span-based localizing network for
  natural language video localization,'' in \emph{ACL}, 2020.

\bibitem{Yang2023DeCoDA}
L.~Yang, Q.~Kong, H.-K. Yang, W.~Kehl, Y.~Sato, and N.~Kobori, ``Deco:
  Decomposition and reconstruction for compositional temporal grounding via
  coarse-to-fine contrastive ranking,'' in \emph{CVPR}, 2023.

\bibitem{Li2023ExploringTE}
C.~Li, Z.~Li, C.~Jing, Y.~Jia, and Y.~Wu, ``Exploring the effect of primitives
  for compositional generalization in vision-and-language,'' in \emph{CVPR},
  2023.

\bibitem{Yuan2018ToFW}
Y.~Yuan, T.~Mei, and W.~Zhu, ``To find where you talk: Temporal sentence
  localization in video with attention based location regression,'' in
  \emph{AAAI}, 2018.

\bibitem{Zeng2020DenseRN}
R.~Zeng, H.~Xu, W.~Huang, P.~Chen, M.~Tan, and C.~Gan, ``Dense regression
  network for video grounding,'' in \emph{CVPR}, 2020.

\bibitem{Lei2021QVHighlightsDM}
J.~Lei, T.~L. Berg, and M.~Bansal, ``Qvhighlights: Detecting moments and
  highlights in videos via natural language queries,'' in \emph{NeurIPS}, 2021.

\bibitem{kingma2014adam}
D.~P. Kingma and J.~Ba, ``Adam: A method for stochastic optimization,'' in
  \emph{ICLR}, 2014.

\end{thebibliography}
\bibliographystyle{IEEEtran}

\end{document}